\documentclass{article}

\usepackage{PRIMEarxiv}

\usepackage[utf8]{inputenc} 
\usepackage[T1]{fontenc}    
\usepackage{hyperref}       
\usepackage{url}            
\usepackage{booktabs}       
\usepackage{amsfonts}       
\usepackage{nicefrac}       
\usepackage{microtype}      
\usepackage{lipsum}
\usepackage{fancyhdr}       
\usepackage{graphicx}       
\usepackage{algorithm}      
\usepackage{algpseudocode}
\usepackage{amsmath}
\usepackage{multirow}
\usepackage{threeparttable}
\graphicspath{{media/}}     

\title{Traversal Learning: A Lossless and Efficient Distributed Learning Framework}

\author{
  Erdenebileg Batbaatar \\
  Neouly Co., Ltd. \\
  South Korea \\
  \texttt{erdenebileg11@gmail.com} \\
  \And
  Jeonggeol Kim \\
  Hongik University \\
  South Korea \\
  \texttt{harin1773@mail.hongik.ac.kr} \\
  \And
  Yongcheol Kim \\
  Neouly Co., Ltd. \\
  South Korea \\
  \texttt{kimycn1017@gmail.com} \\
  \And
  Young Yoon \\
  Hongik University \\
  South Korea \\
  \texttt{young.yoon@hongik.ac.kr} \\
}

\begin{document}
\maketitle

\begin{abstract}
In this paper, we introduce Traversal Learning (TL), a novel approach designed to address the problem of decreased quality encountered in popular distributed learning (DL) paradigms such as Federated Learning (FL), Split Learning (SL), and SplitFed Learning (SFL). Traditional FL experiences from an accuracy drop during aggregation due to its averaging function, while SL and SFL face increased loss due to the independent gradient updates on each split network. TL adopts a unique strategy where the model traverses the nodes during forward propagation (FP) and performs backward propagation (BP) on the orchestrator, effectively implementing centralized learning (CL) principles within a distributed environment. The orchestrator is tasked with generating virtual batches and planning the sequential node visits of the model during FP, aligning them with the ordered index of the data within these batches. We conducted experiments on six datasets representing diverse characteristics across various domains. Our evaluation demonstrates that TL is on par with classic CL approaches in terms of accurate inference, thereby offering a viable and robust solution for DL tasks. TL outperformed other DL methods and improved accuracy by 7.85\% for independent and identically distributed (IID) datasets, macro F1-score by 1.06\% for non-IID datasets, accuracy by 2.60\% for text classification, and AUC by 3.88\% and 4.54\% for medical and financial datasets, respectively. By effectively preserving data privacy while maintaining performance, TL represents a significant advancement in DL methodologies. The implementation of TL is available at \url{https://github.com/neouly-inc/Traversal-Learning}.
\end{abstract}


\begingroup
    \section{Introduction}\label{sec:introduction}

While distributed learning (DL) methods offer scalability and privacy benefits \cite{cao2023communication,dehghani2023distributed,liu2022wireless,froelicher2020scalable}, they can also encounter challenges related to quality loss \cite{liang2024communication,aach2023large,rojas2021large}. Factors such as resource constraints \cite{an2023towards,kirienko2021distributed}, privacy-preserving techniques \cite{alshammari2024privacy,mohammadi2021differential}, data heterogeneity \cite{an2023towards,kirienko2021distributed}, and synchronization issues \cite{kirienko2021distributed,alshammari2024privacy} can lead to accuracy loss in DL methods. In general, there are two types of DL methods: (1) Parallel learning methods, such as Federated Learning (FL) \cite{kairouz2021advances}, distribute model training across decentralized devices (also referred to as \textit{nodes}), enabling privacy-preserving training on local data while generating a global model through periodic aggregation of model updates. The aggregation inevitably causes accuracy loss \cite{tang2023fedrad,gao2023high,guo2022adaptive,xu2022lazy}; (2) In contrast, sequential DL methods, such as Split Learning (SL) \cite{gupta2018distributed,vepakomma2018split}, segmenting the model architecture, with data flowing sequentially through these segments \cite{vepakomma2022split}. Splitting the model across different devices for privacy reasons and conducting separate gradient calculations on each node can lead to accuracy loss \cite{abedi2024fedsl,han2021accelerating}. Addressing these challenges is crucial to ensuring the effectiveness of DL approaches in preserving data privacy without compromising accuracy compared to methods that aggregate training data in a central location \cite{bouramoul2023enhancing,pham2022split}.

In DL methods, local data from multiple nodes is utilized collaboratively without sharing raw data, ensuring privacy. However, challenges such as data heterogeneity and data imbalance can lead to a catastrophic forgetting problem \cite{babakniya2024data,wei2022knowledge}. To deal with this issue, we propose an entity referred to as the \textit{orchestrator} that first separates the forward propagation (FP) concerns to the nodes. The orchestrator performs global backward propagation (BP) based on the first-layer activations, first-layer gradients, and last-layers gradients collected from the nodes, thereby ensuring consistent model parameter updates throughout the distributed system. To reduce runtime and improve efficiency, data transfers between the orchestrator and the nodes are performed in parallel, minimizing communication latency and improving the overall performance. The orchestrator maintains \textit{virtual batches} with data indices of the local samples in a random order. The virtual batches are used to determine the node traversal during the FP phase. The virtual batches play a key role in privacy-preserving collaboration between the orchestrator and the nodes. We refer to this new DL scheme as Traversal Learning (TL), effectively realizes the classical centralized learning (CL) principles in a distributed environment, which benefits application domains such as healthcare and finance, where highly sensitive data must not be shared and quality degradation cannot be tolerated. To the best of our knowledge, these ideas have not been explored in other works. Furthermore, this work opens up many intriguing research challenges in the DL realm.
    \section{Related Works}\label{sec:related}

We situate our work within the context of DL architectures and efforts to address quality degradation issues.

In FL, a server updates a global model by securely aggregating local models that are independently trained by distributed clients with local data not shared with others \cite{bejenar2023aggregation,ntantiso2022review,mansour2022federated}. FL represents a seminal privacy-preserving DL method. Several variations of FL have been developed, including Federated Averaging (FedAvg) \cite{sun2022decentralized,deng2020distributionally,wang2020federated}, FedProx \cite{su2023non,yuan2022convergence,li2020federated}, Personalized FL \cite{tan2022towards,deng2020adaptive,t2020personalized}, and Federated Transfer Learning \cite{saha2021federated,liu2020secure,chen2020fedhealth}. However, the fundamental nature of FL, which relies on averaging local models, limits the robustness of the global model compared to centralized learning approaches that have full access to the training data. Whereas FL aggregates local models in parallel, SL nodes and the server sequentially update horizontally and vertically partitioned components of the model. In vanilla SL, each client device trains its predetermined portion of the model locally without directly accessing the labels and sends its 'smashed data' to the server, which computes the loss using the server's portion of the model. To address privacy concerns, SL without label sharing (SL+), has been introduced, in which the initial and final portions of the model are kept on the client side, and only the middle portion is shared with the server \cite{gupta2018distributed,vepakomma2018split}. 

FL aggregation is employed to merge the updates of split model parts in Split Federated Learning (SFL) \cite{thapa2022splitfed}. Merging the independent gradient updates of split parts in SL and SFL results in higher loss compared to CL approaches \cite{pal2021server}. A key reason for the increased loss in SFL and SL setups is data heterogeneity—the non-independent and identically distributed (non-IID) nature of client data distributions. Since clients train their portions of the model on different data distributions, gradient aggregation can be suboptimal, causing the global model to converge more slowly or less effectively. This issue is more pronounced compared to CL, where the model has access to the full dataset at once and can optimize without needing to aggregate partial updates. SFL continues to face robustness limitations, particularly in environments with significant data variability across nodes \cite{liao2024mergesfl,yao2019federated,xu2024multi}.

Several recent works studied the accuracy loss issue inherent in existing DL methods. One of the main reasons for accuracy loss in DL is data heterogeneity across clients, which can result in client drift and prevent the global model from effectively representing the overall data distribution. This problem is well studied in \cite{karimireddy2019scaffold}, where the authors introduced the SCAFFOLD algorithm to mitigate this issue. Additionally, non-IID settings were analyzed, and improvements in client selection and adaptive aggregation strategies were proposed to mitigate accuracy loss \cite{li2024comprehensive}. Model fusion techniques in DL often lead to generalization errors due to overfitting on local client datasets. Ji et al. \cite{ji2102emerging} have explored how these errors arise in FL setups and highlighted challenges with aggregation of diverse local models. They show that model fusion in FL and similar distributed methods struggle to generalize as effectively as CL. In addition, Pillutla et al. \cite{pillutla2022robust} have explored robust aggregation techniques to alleviate some of these issues, but generalization remains a challenge. Communication delays and asynchronous updates in DL environments can cause stale updates to be incorporated into the global model, which degrades accuracy. Xie et al. \cite{xie2019asynchronous} recently studied asynchronous federated learning (FedAsync) and demonstrated that, while asynchronous updates improve scalability, they negatively affect the model’s accuracy due to outdated information. Most recently, Lu et al. \cite{lu2024adaptive} proposed a new asynchronous update scheme that reduces accuracy loss while maintaining scalability. In SL and SFL frameworks, the server aggregates partial model updates from clients, which can lead to suboptimal global models due to incomplete information. Thapa et al. \cite{thapa2022splitfed} highlighted the greater loss associated with SL setups compared to centralized learning, particularly due to poor gradient aggregation. Recent work by Shiranthika et al. \cite{shiranthika2024optimizing} introduced an improved aggregation method in SplitFed, but the fundamental limitations still remain. Unlike centralized models, distributed models do not have access to the full dataset, limiting their ability to learn comprehensive patterns across all data points. Mao et al. \cite{mao2023safari} discussed how FL suffers from this limitation, especially when dealing with sparse or biased datasets. Babar et al. \cite{babar2024investigating} also analyzed how FL models perform worse due to the limited data representation inherent in distributed settings. FedAvg-based algorithms apply implicit regularization by averaging model updates, which helps reduce overfitting in some cases but can also result in a weaker global model in terms of accuracy. Su et al. \cite{su2023non} discussed how this regularization leads to performance degradation in highly heterogeneous environments. FedProx~\cite{li2020federated} was designed to address this issue but still struggles in scenarios with extreme data imbalance. As local models are trained on client-specific data, their updates can drift significantly, leading to less representative global models. FedCSD~\cite{yan2023rethinking} uses a class prototype similarity distillation method to mitigate drift issues. Tan et al. \cite{tan2022towards} also demonstrated how personalized FL techniques can reduce client drift, albeit at the cost of increased complexity.

Existing DL architectures that rely on model fusion are inherently susceptible to generalization errors \cite{ji2024emerging}, which limits their ability to perform on par with CL. This challenge has prompted a significant shift from traditional DL models to a novel architecture referred to as TL. In TL, local nodes contribute to the FP of a global model in a synchronized manner, while the server performs BP and model optimization after each batch. TL effectively implements the principles of CL in a decentralized setting, enabling it to perform as competitively as CL approaches without depending on the quality enhancement techniques explored in previous studies. In the following section, we introduce the key mechanisms that ensure TL achieves a quality comparable to that of CL methods.

In the rest of this paper, we present the details of the TL functionalities and analyze its unique non-functional properties through experiments with diverse public datasets to evaluate its real-world relevance and applicability.
    \section{Traversal Learning}\label{sec:method}

TL is a novel DL paradigm that combines the strengths of CL with the privacy-preserving features of DL methods. TL achieves this by coordinating FP across distributed nodes and performing BP on a central orchestrator. This section outlines the fundamental components of TL, focusing on (1) virtual batch creation, (2) training workflow, (3) FP and BP in a distributed environment, and (4) mechanisms for synchronization and communication efficiency. These integrated features establish TL as a scalable and reliable framework, ideally suited for addressing the challenges of modern DL applications, particularly in domains requiring stringent data privacy. Each of these features is detailed in the following sections.

\begin{figure}[htbp]
    \centerline{\includegraphics[width=0.5\linewidth]{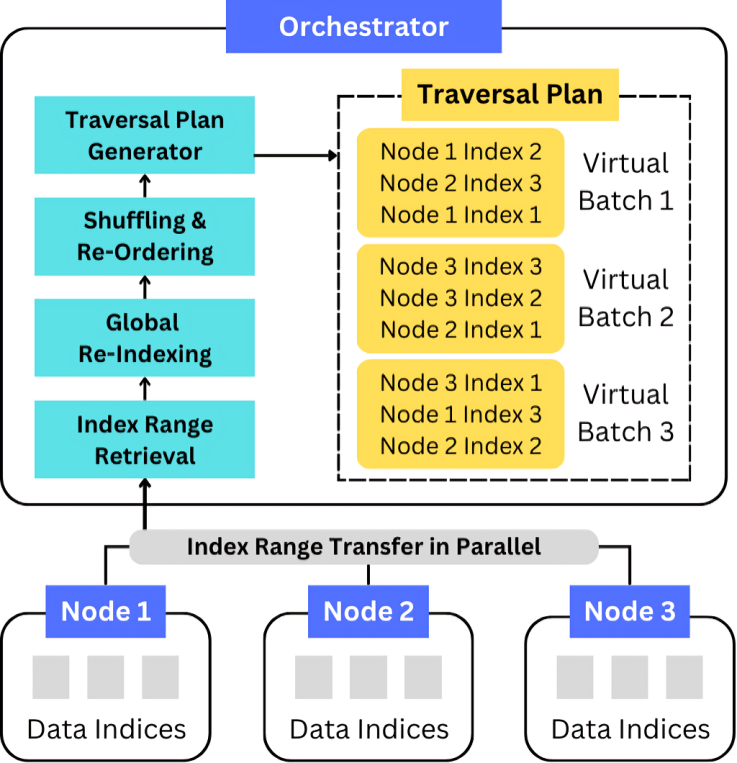}}
    \caption{Virtual batch creation scheme shows how the TL orchestrator retrieves data indices, performs global re-indexing, creates shuffled virtual batches, and generates a traversal plan for efficient node traversal during FP.}
    \label{fig1}
\end{figure}

\subsection{Virtual Batch Creation}\label{subsec:virtual-batch-creation}

Virtual batch creation is a critical feature of TL, enabling the orchestrator to manage and optimize data flow across distributed nodes, as shown in Figure \ref{fig1}. It ensures efficient traversal of nodes during FP and guarantees consistency in BP. This mechanism involves the following key steps; such as index range retrieval, global re-indexing, shuffling and re-ordering, and traversal plan generation; which are detailed in Algorithm \ref{alg1}. These steps are designed to enhance model generalization, maintain synchronization, and minimize communication overhead.

\begin{enumerate}

    \item \textit{Index Range Retrieval}. The first step in virtual batch creation is index range retrieval, which establishes a clear understanding of how data is distributed across nodes in the TL framework. Each node independently indexes its local dataset, and the orchestrator queries all participating nodes to collect these index ranges. For instance, a node with 100 samples might have an index range of [0, 99]. These ranges allow the orchestrator to construct a global map of the dataset’s structure without accessing raw data, preserving privacy. This mapping provides a blueprint for tracking data allocation across nodes and serves as the foundation for subsequent steps, such as global re-indexing and virtual batch creation. By consolidating these index ranges accurately, the orchestrator ensures that each node’s data is represented in the training process, setting the stage for efficient and synchronized FP and BP.
    
    \item \textit{Global Re‐Indexing}. After gathering the index ranges, the orchestrator assigns a unique global identifier to each data point, creating a cohesive global mapping of the dataset. This holistic view ensures that all data points are systematically incorporated into training, avoiding overfitting to local distributions or catastrophic forgetting. Global re-indexing is especially critical in non-IID settings, where data heterogeneity across nodes can lead to biased or suboptimal learning. By integrating data points from all nodes, the orchestrator prepares the dataset for virtual batch creation and traversal planning. This structured global map enhances model generalization, ensures fair representation of all data, and supports efficient synchronization during training while upholding data privacy.
    
    \item \textit{Shuffling and Re‐Ordering}. To enhance model generalization and mitigate biases, the orchestrator shuffles and reorders the global index map after re-indexing. This step randomizes the order in which data points are processed, reducing the risk of overfitting to localized patterns. The shuffled indices are grouped into virtual batches, combining samples from multiple nodes to ensure a balanced and diverse representation during training. This randomized approach is particularly effective in non-IID scenarios, where data heterogeneity can otherwise degrade model performance. By fostering robustness and reducing bias, shuffling and re-ordering further support TL’s privacy-preserving and generalization goals.
    
    \item \textit{Traversal Plan Generation}. The final step, traversal plan generation, ensures efficient coordination of FP across distributed nodes. Using the shuffled virtual batches, the orchestrator constructs a detailed plan dictating the sequence of node visits during training. This plan integrates diverse data points from multiple nodes, promoting balanced generalization while minimizing communication overhead through optimized data transfers and reduced idle time. By synchronizing the contributions of all nodes, the traversal plan maintains consistency and avoids delays or inconsistencies that might degrade model quality. Designed to scale seamlessly with an increasing number of nodes, this plan ensures the efficiency and robustness of TL, enabling it to emulate centralized learning principles within a distributed environment.
    
\end{enumerate}

\begin{algorithm}
    \caption{Virtual Batch Creation in Traversal Learning.}
    \label{alg1}
    \begin{algorithmic}[1]
    \State \textbf{Input:} Local datasets $\mathcal{D}_1, \mathcal{D}_2, \ldots, \mathcal{D}_n$ on nodes $N_1, N_2, \ldots, N_n$
    \State \textbf{Output:} Global index map and virtual batches for training
    \State \textbf{Initialization:} Orchestrator $\mathcal{O}$ queries nodes for index ranges
    
    \Procedure{Index Range Retrieval}{}
        \For{each node $N_i$}
            \State Retrieve local index range $\mathcal{I}_i$ from $\mathcal{D}_i$
            \State Send $\mathcal{I}_i$ to orchestrator $\mathcal{O}$
        \EndFor
    \EndProcedure
    
    \Procedure{Global Re-Indexing}{}
        \State $\mathcal{O}$ constructs global index map $\mathcal{I}_{global}$ by assigning a unique global index to each data point in all nodes' datasets $\mathcal{D}_1, \mathcal{D}_2, \ldots, \mathcal{D}_n$
    \EndProcedure
    
    \Procedure{Shuffling and Re-Ordering}{}
        \State Shuffle the global index map $\mathcal{I}_{global}$ to create randomized virtual batches $\mathcal{B}_1, \mathcal{B}_2, \ldots, \mathcal{B}_m$
    \EndProcedure
    
    \Procedure{Traversal Plan Generator}{}
        \State Generate traversal plan for nodes based on the virtual batches $\mathcal{B}_1, \mathcal{B}_2, \ldots, \mathcal{B}_m$
        \State Each batch $\mathcal{B}_j$ defines the sequence of nodes to visit during FP
    \EndProcedure
    
    \State \textbf{Return:} Traversal plan and virtual batches $\mathcal{B}_1, \mathcal{B}_2, \ldots, \mathcal{B}_m$
    \end{algorithmic}
\end{algorithm}

\subsection{Training Procedure}\label{subsec:training-procedure}

The training procedure in TL is meticulously designed to coordinate FP and BP across distributed nodes, ensuring synchronization, efficiency, and high model performance, as shown in Figure \ref{fig2}. This process is centrally orchestrated and involves several critical steps, which are outlined in detail in Algorithm \ref{alg2}:

\begin{itemize}
    \item \textit{Traversal Scheduling}. The orchestrator initiates the training process by implementing the traversal plan generated during the virtual batch creation phase. This scheduler determines the sequence in which nodes are visited and assigns specific subsets of data from the virtual batches to each node. During FP, the model traverses through the nodes in the specified order, with each node processing its allocated data. This step ensures balanced participation of all nodes, enabling an efficient and privacy-preserving training process.
    \item \textit{Activation and Gradient Retrieval}. After the FP phase, the orchestrator collects the first-layer activations, first-layer gradients, and last-layer gradients from all participating nodes. This step minimizes communication overhead by transferring only essential gradient and activation information instead of the entire model or activations from deeper layers. The orchestrator then aggregates these values to perform global backward propagation, ensuring consistency and alignment across the distributed system.
    \item \textit{Centralized Backward Propagation}. Using the aggregated first-layer activations, first-layer gradients, and last-layer gradients, the orchestrator performs BP to update the model parameters. This centralized approach addresses common challenges in decentralized learning methods, such as model drift and inconsistent gradient updates, ensuring synchronized parameter optimization. By keeping BP centralized, TL achieves a level of accuracy and robustness comparable to CL, even in a distributed setting.
    \item \textit{Model Redistribution}. Once BP is complete, the orchestrator redistributes the updated model to all nodes, enabling the next iteration of FP. This iterative loop continues until the model converges, with each cycle leveraging the strengths of both distributed data processing and centralized parameter optimization.
    \item \textit{Parallelized Communication}. Throughout the training process, TL optimizes communication efficiency by enabling parallel data transfers between the nodes and the orchestrator. While one node completes its FP task, the orchestrator prepares the next node to begin processing. This pipelined approach reduces idle time and ensures that all nodes contribute seamlessly to the training process.
\end{itemize}

\begin{figure}[htbp]
    \centerline{\includegraphics[width=0.5\linewidth]{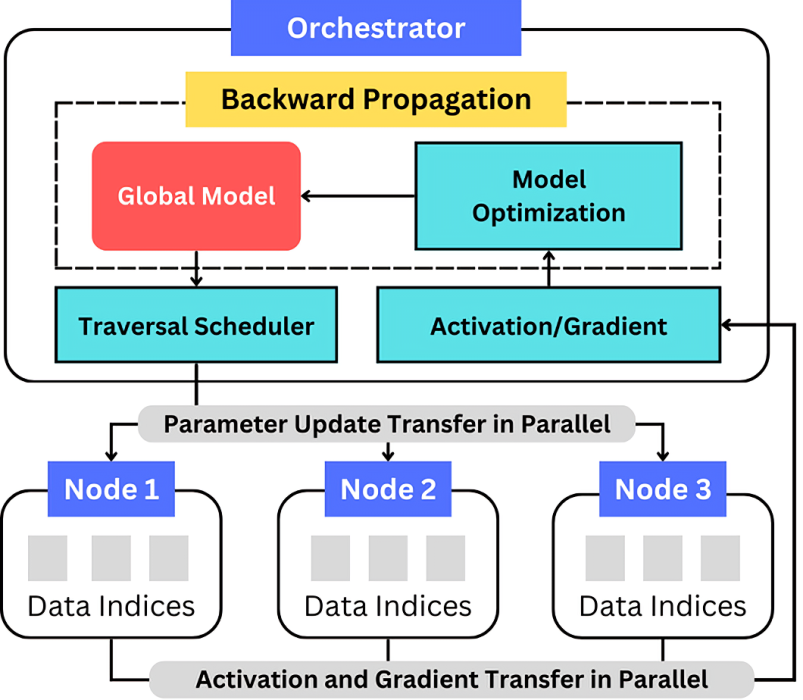}}
    \caption{Training procedure in TL illustrates how the orchestrator manages FP and BP, scheduling node visits, collecting first-layer activations and last-layer gradients, recalculating subsequent layer activations using model parameters, and performing centralized BP starting from the last-layer gradients for consistent model training.}
    \label{fig2}
\end{figure}

\begin{algorithm}
    \caption{Training Procedure in Traversal Learning}
    \label{alg2}
    \begin{algorithmic}[1]
        \State \textbf{Input:} Virtual batches $\mathcal{B}_1, \mathcal{B}_2, \ldots, \mathcal{B}_m$, model $\mathcal{M}$, orchestrator $\mathcal{O}$, nodes $N_1, N_2, \ldots, N_n$
        \State \textbf{Output:} Trained model $\mathcal{M}$
        
        \Procedure{Traversal Scheduler}{}
            \For{each virtual batch $\mathcal{B}_j$}
                \For{each node $N_i$ in the traversal plan for $\mathcal{B}_j$}
                    \State Send model $\mathcal{M}$ to node $N_i$
                    \State Node $N_i$ performs FP on data subset from $\mathcal{B}_j$
                \EndFor
            \EndFor
        \EndProcedure
        
        \Procedure{Activation and Gradient Retrieval}{}
            \For{each virtual batch $\mathcal{B}_j$}
                \State Orchestrator $\mathcal{O}$ collects first-layer activations $X^{(1)}_i$, first-layer gradients $\frac{\partial \mathcal{L}^{(i)}}{\partial X^{(1)}_i}$, and last-layer gradients $\delta^{(L)}_i$ from nodes $N_1, N_2, \ldots, N_n$
                \State Aggregate these values for global BP
            \EndFor
        \EndProcedure
        
        \Procedure{Model Optimization}{}
            \State Orchestrator $\mathcal{O}$ recalculates activations for all layers using aggregated $X^{(1)}_i$ and model parameters
            \State Orchestrator $\mathcal{O}$ performs BP starting from aggregated $\delta^{(L)}_i$ using recalculated activations
            \State Update model parameters $\theta$ on the orchestrator
            \State Send updated model $\mathcal{M}$ back to nodes
        \EndProcedure
        
        \State \textbf{Return:} Trained model $\mathcal{M}$
    \end{algorithmic}
\end{algorithm}

Batch processing in TL is a pivotal mechanism enabling efficient handling of large datasets across distributed nodes. It ensures streamlined training by dividing the global dataset into smaller, shuffled, and reordered virtual batches. This segmentation balances computational loads among nodes, enhances parallel processing, and prevents biases. During FP, nodes process their respective portions of a virtual batch sequentially based on a traversal plan, reducing computational strain and improving operational parallelization. The orchestrator oversees the coordination of FP and BP. After collecting first-layer activations and last-layer gradients post-FP from the nodes, the orchestrator recalculates the activations for all subsequent layers using the first-layer activations and current model parameters, then performs centralized BP starting from the aggregated last-layer gradients. This method reduces unnecessary data transfers and minimizes latency by selectively updating only involved nodes with revised model parameters. TL's batch processing framework ensures synchronized and efficient model training while optimizing system performance in large-scale distributed environments.

\subsection{Hybrid Framework}\label{subsec:forward-and-backward-propagation-distributed-environment}

TL employs a hybrid framework that synergistically combines the benefits of decentralized and centralized learning paradigms. FP is distributed across multiple nodes, allowing local data processing to occur where the data resides, while BP is centralized on an orchestrator to optimize model parameters efficiently. This strategic framework is carefully designed to address the inherent challenges of DL systems, such as communication overhead, scalability, and data privacy, making TL a robust and efficient solution for diverse applications. The hybrid framework leverages distributed FP to ensure that raw data remains on the nodes, thereby maintaining privacy and adhering to strict data regulations in sensitive domains such as healthcare and finance. Simultaneously, centralizing BP enables consistent and synchronized model parameter updates, a key factor in mitigating the challenges of model drift and divergence often encountered in fully decentralized approaches. By implementing this hybrid design, TL ensures that both scalability and data security are achieved without compromising the accuracy and robustness of the global model. Below, the detailed processes for distributed and centralized learning in TL are elaborated:

\subsubsection{Distributed Phase}

In TL, FP is distributed across nodes, with each node processing a portion of the global dataset independently. The steps involved include:


    
    
    

    



\begin{enumerate}

    \item \textit{Local Data Processing}. Each node computes the first-layer activations and gradients for its local data subset. Let \( X^{(i)} \) represent the input data on node \( i \), and \( W^{(1)} \), \( b^{(1)} \) denote the weights and biases of the first layer. The pre-activation output \( Z^{(1)}_i \) is calculated as:
    
    \begin{equation}
        Z^{(1)}_i = W^{(1)} X^{(i)} + b^{(1)}
    \end{equation}
    
    The activation function \( f \), such as Rectified Linear Unit (ReLU) or Sigmoid, is then applied to produce the first-layer activations:
    
    \begin{equation}
        X^{(1)}_i = f(Z^{(1)}_i)
    \end{equation}
    
    The node also computes the first-layer gradient \( \frac{\partial \mathcal{L}^{(i)}}{\partial X^{(1)}_i} \) during local backward propagation, based on the downstream gradients received from subsequent layers.

    \item \textit{Last-Layer Gradient Calculation}. Each node performs forward propagation through the full model locally to compute the predicted outputs \( \hat{y}^{(i)} \) and true labels \( y^{(i)} \), then calculates the last-layer gradient \( \delta^{(L)}_i \) as:
    
    \begin{equation}
        \delta^{(L)}_i = \frac{\partial \mathcal{L}^{(i)}}{\partial \hat{y}^{(i)}} \otimes f^{\prime}(Z^{(L)}_i)
    \end{equation}
    
    where \( \mathcal{L}^{(i)} = \mathcal{L}(\hat{y}^{(i)}, y^{(i)}) \) is the local loss, and \( Z^{(L)}_i \) is the pre-activation output of the last layer.

    \item \textit{Transmission to the Orchestrator}. Nodes send their first-layer activations \( X^{(1)}_i \), first-layer gradients \( \frac{\partial \mathcal{L}^{(i)}}{\partial X^{(1)}_i} \), and last-layer gradients \( \delta^{(L)}_i \) to the orchestrator. By transmitting only these specific values, TL minimizes communication overhead compared to transferring the entire model or activations and gradients from all layers.

\end{enumerate}

\subsubsection{Centralized Phase}

BP is centralized at the orchestrator, ensuring consistent and synchronized updates to the model parameters. The process includes:

\begin{enumerate}

    \item \textit{Activation Recalculation}. The orchestrator recalculates the activations for all layers using the aggregated first-layer activations \( X^{(1)}_i \) from all nodes and the current model parameters. For each layer \( l > 1 \), the pre-activation output \( Z^{(l)}_i \) and activation \( X^{(l)}_i \) are computed as:
    
    \begin{equation}
        Z^{(l)}_i = W^{(l)} X^{(l-1)}_i + b^{(l)}
    \end{equation}
    
    \begin{equation}
        X^{(l)}_i = f(Z^{(l)}_i)
    \end{equation}

    \item \textit{Gradient Aggregation and Calculation}. The orchestrator aggregates the last-layer gradients \( \delta^{(L)}_i \) and first-layer gradients \( \frac{\partial \mathcal{L}^{(i)}}{\partial X^{(1)}_i} \) from all nodes, then calculates the gradients for all model parameters using the chain rule and recalculated activations:
    
    \begin{itemize}
        \item The aggregated last-layer gradient is:
        \begin{equation}
            \delta^{(L)} = \frac{1}{n} \sum_{i=1}^{n} \delta^{(L)}_i
        \end{equation}
        
        \item The gradients for the last-layer weights \( W^{(L)} \) and biases \( b^{(L)} \) are:
        \begin{equation}
            \frac{\partial \mathcal{L}_{global}}{\partial W^{(L)}} = \delta^{(L)} (X^{(L-1)})^T
        \end{equation}
        
        \begin{equation}
            \frac{\partial \mathcal{L}_{global}}{\partial b^{(L)}} = \delta^{(L)}
        \end{equation}
        
        \item For intermediate layers \( l \) (from \( L-1 \) to 2), gradients are computed by backpropagating \( \delta^{(l+1)} \) through the recalculated activations:
        \begin{equation}
            \delta^{(l)} = (W^{(l+1)})^T \delta^{(l+1)} \odot f'(Z^{(l)})
        \end{equation}
        \begin{equation}
            \frac{\partial \mathcal{L}_{global}}{\partial W^{(l)}} = \delta^{(l)} (X^{(l-1)})^T
        \end{equation}
        \begin{equation}
            \frac{\partial \mathcal{L}_{global}}{\partial b^{(l)}} = \delta^{(l)}
        \end{equation}
        
        \item For the first layer, the aggregated first-layer gradient is:
        \begin{equation}
            \frac{\partial \mathcal{L}_{global}}{\partial X^{(1)}} = \frac{1}{n} \sum_{i=1}^{n} \frac{\partial \mathcal{L}^{(i)}}{\partial X^{(1)}_i}
        \end{equation}
        ensuring consistency with the recalculated forward pass.
    \end{itemize}

    \item \textit{Parameter Updates}. The orchestrator updates the model parameters using gradient descent. For each layer \( l \), with learning rate \( \eta \), the weights and biases are updated as:
    
    \begin{equation}
        W^{(l)} \leftarrow W^{(l)} - \eta \frac{\partial \mathcal{L}_{global}}{\partial W^{(l)}}
    \end{equation}
    
    \begin{equation}
        b^{(l)} \leftarrow b^{(l)} - \eta \frac{\partial \mathcal{L}_{global}}{\partial b^{(l)}}
    \end{equation}

    \item \textit{Redistribution to Nodes}. The updated model parameters are redistributed to the nodes, allowing them to initiate the next round of FP with improved parameters.
        
\end{enumerate}

\subsection{Synchronization and Communication}\label{subsec:synchronization-and-communication-efficiency}

Efficient synchronization and communication are critical for the success of DL architectures. These components ensure the coordination of operations between nodes and the orchestrator, maintaining consistency, minimizing latency, and optimizing resource utilization. TL employs innovative techniques to address challenges such as communication overhead, synchronization delays, and system scalability, ensuring robust and efficient training across distributed systems.

\textbf{Synchronization Mechanisms}: Synchronization is essential in distributed frameworks to ensure that all nodes contribute to the global model consistently and effectively. In TL, the orchestrator plays a central role in maintaining this synchronization by coordinating FP and BP phases. The orchestrator generates a detailed traversal plan based on the virtual batches. This plan dictates the sequence in which nodes process data during FP, ensuring that every node contributes equitably to the training process. While one node completes its FP task, the orchestrator prepares subsequent nodes for processing, reducing idle time. This pipelined approach optimizes resource utilization and ensures a seamless flow of tasks across the distributed network. After each BP phase, the orchestrator synchronizes model updates across all nodes, ensuring that every node operates with the most recent global parameters in the next iteration. This minimizes the risk of divergence and maintains the integrity of the learning process.

\textbf{Communication Efficiency}: Communication overhead is a major bottleneck in distributed learning (DL). TL addresses this challenge by implementing strategies that reduce data transmission requirements while maintaining the quality of model updates. Instead of transmitting entire models or intermediate activations from all layers, TL restricts communication to first-layer activations, first-layer gradients, and last-layer gradients. This significantly reduces the amount of data exchanged between nodes and the orchestrator, enhancing communication efficiency. TL enables concurrent data exchanges between nodes and the orchestrator. For example, while one node sends its first-layer activations and gradients, another node can begin its FP phase. This parallelization reduces overall training time and ensures continuous operation across the system. During BP, the orchestrator uses the collected last-layer gradients to initiate centralized updates, recalculating subsequent layer activations from the first-layer activations and model parameters, thus avoiding the need to transmit additional intermediate gradients. Unimportant gradients can be compressed or omitted based on their relevance to global optimization, further reducing bandwidth usage without compromising model accuracy. TL employs data compression methods to minimize the size of transmitted information. By compressing first-layer activations, first-layer gradients, and last-layer gradients, TL further reduces communication overhead, making it suitable for resource-constrained environments.

\textbf{Asynchronous Updates}: In real-world distributed systems, network delays, node failures, and asynchronous updates can disrupt synchronization and degrade model performance. TL mitigates these issues through the following measures. The orchestrator stores delayed updates from nodes in a gradient buffer and aggregates them only when sufficient updates are received. This prevents stale gradients from negatively impacting the global model. TL adapts its traversal schedule in response to network conditions and node availability. By prioritizing nodes with faster updates, TL reduces the impact of delays and ensures efficient model updates. In high-latency environments, TL adjusts the frequency of synchronization to balance accuracy and communication costs. Nodes may perform multiple FP passes before synchronizing with the orchestrator, conserving bandwidth while maintaining model quality.

\textbf{Trade-offs and Optimization}: TL recognizes the inherent trade-offs between synchronization and communication efficiency. While strict synchronization ensures high accuracy and consistency, it can increase communication delays in bandwidth-constrained systems. Conversely, looser synchronization policies improve scalability but may introduce minor accuracy degradation. TL offers configurable synchronization parameters, allowing users to balance these trade-offs based on the specific requirements of their applications. For critical tasks like medical diagnostics, TL enforces strict synchronization to ensure every node’s updates are incorporated in real-time. In resource-constrained settings, TL allows for reduced synchronization frequency, enabling efficient training with minimal communication overhead.
    \section{Evaluation}\label{sec:evaluation}

This section presents an evaluation of TL in terms of performance, runtime, and scalability. We compare TL against CL and other distributed learning methods, including FL, SL, SL+, and SFL. The evaluation encompasses diverse datasets and learning models to assess TL’s effectiveness under various conditions.

\subsection{Experimental Setup}\label{subsec:experimental-setup}

We conducted our experiments on a server with a 40-core CPU with 80 threads, 256GB of RAM, and six Tesla V100 GPUs, each with 32GB of dedicated memory. This server is operated on Ubuntu 18.04.6 LTS, and the DL jobs were executed on Docker version 24.0.2, with Ubuntu 22.04.2 LTS containers, using full access to memory via the inter-process computation option and access to the GPU devices on the server. All our deep learning algorithms were implemented with
TensorFlow 2.13.0 \cite{abadi2016tensorflow} in Python 3.11. Additionally, all experiments were conducted in a simulation environment to test and validate TL under controlled conditions.

\subsubsection{Datasets}

For our experiment, we used various types of datasets from diverse domains to investigate the effect of data distribution. For well-balanced IID setting, we used popular image datasets such as MNIST \cite{deng2012mnist} and CIFAR-10 \cite{krizhevsky2009learning}. In contrast, the NICO \cite{he2021towards} dataset was used for a non-IID setting. We also considered privacy-sensitive domains such as medical and financial sectors to apply the DL methods. Both domains provided imbalanced binary IID datasets. Specifically, we used the MIMIC-IV \cite{xie2022benchmarking} dataset from the medical domain and Bank Marketing \cite{moro2014data} dataset from the financial domain. To simulate non-IID datasets among nodes, we applied K-Means clustering \cite{ahmed2020k} to partition the MIMIC-IV and Bank Marketing datasets into multiple subsets. Each subset was then assigned to a different node. Additionally, for text classification tasks, we employed the IMDB movie reviews dataset \cite{maas-EtAl:2011:ACL-HLT2011}, a balanced binary IID dataset used for sentiment analysis. We provide a detailed overview of each dataset below, highlighting their characteristics and relevance to the experimental evaluation:

\begin{itemize}

    \item \textit{MNIST}. A widely used dataset in machine learning, consisting of 28x28 grayscale images of handwritten digits (0-9). It is often used for training and testing image classification algorithms.
    
    \item \textit{CIFAR-10}. Another popular image dataset consisting of 60,000 images, each with a size of 32x32 pixels, divided into 10 classes, with 6,000 images per class. It is used for evaluating image classification models.
    
    \item \textit{NICO}. This non-IID dataset simulates differences between training and testing distributions. NICO includes two superclasses, animal and vehicle, with 19 classes and about 25,000 images. This dataset is designed for non-IID image classification tasks, simulating scenarios where testing distributions differ significantly from training distributions. It includes images labeled with both primary concepts and contextual settings, facilitating research in transfer learning, domain adaptation, stable learning, and domain generalization. NICO has two superclass categories: animal and vehicle, comprising 19 classes and nearly 25,000 images.
    
    \item \textit{MIMIC-IV}. A large, publicly available health record dataset contains de-identified data from critical care patients, including demographics, vital signs, lab tests, medications, and outcomes. MIMIC-IV is used for studies in predictive modeling, clinical decision support, and patient monitoring. A large, publicly available electronic health record dataset containing de-identified health data from patients admitted to critical care units. It includes information such as demographics, vital signs, laboratory tests, medications, and outcomes. Researchers use MIMIC-IV for various healthcare-related studies, including predictive modeling, clinical decision support, and patient monitoring.
    
    \item \textit{BANK}. This dataset contains data from direct marketing campaigns of a Portuguese bank, including client demographics, contact history, and campaign outcomes. It is used to develop models predicting customer responses to marketing efforts. This dataset contains information related to the direct marketing campaigns of a Portuguese banking institution. It includes features such as client demographics, contact history, and campaign outcomes (e.g., client subscription to a term deposit). BANK is used to develop models for predicting customer responses to marketing campaigns.
    
    \item \textit{IMDB}. This dataset of 50,000 movie reviews, labeled as positive or negative, is widely used for sentiment analysis and evaluating Natural Language Processing (NLP) models, particularly in binary sentiment classification. Its varied review lengths make it valuable for testing text-based deep learning models. A widely-used text dataset for sentiment analysis, consisting of 50,000 movie reviews labeled as either positive or negative, with a balanced distribution of sentiment classes. It is used to evaluate natural language processing (NLP) models, particularly in binary sentiment classification tasks. The reviews vary in length and content, making it a valuable dataset for testing the performance of text-based deep learning models.

\end{itemize}

\subsubsection{Deep Learning Models} 

We trained a range of popular deep learning models for our experiments as follows: ResNet-18 \cite{he2016deep} for MNIST, LeNet-5 \cite{lecun1998gradient} for CIFAR-10, ConvNet \cite{he2021towards} for NICO, Datret \cite{githubAbdualimovTPdatret} for both MIMIC-IV and BANK, and Transformer \cite{vaswani2017attention} for IMDB.

\begin{itemize}

    \item \textit{ResNet-18}. ResNet-18 is a convolutional neural network designed for robust image classification using residual blocks to improve gradient flow and prevent vanishing gradients. Our implementation starts with a convolutional layer (64 filters), followed by batch normalization, ReLU activation, and max-pooling. Four stages of residual blocks with increasing filter sizes (64, 128, 256, 512) are used, with downsampling through strides of 2. A global average pooling layer and fully connected softmax layer complete the model for effective classification.

    \item \textit{LeNet-5}. LeNet-5 is a classic convolutional neural network for image classification. Our implementation includes two convolutional layers (6 filters, then 16 filters), both followed by max-pooling and using the 'swish' activation function. The output is flattened, with dropout applied to prevent overfitting. Two fully connected layers (120 and 84 units) are also followed by dropout. The final output layer is a softmax-activated dense layer for classification.

    \item \textit{ConvNet}. ConvNet was used for complex image classification tasks, employing five convolutional layers with progressively increasing filters (64, 128, 256, 512, 1024). Each layer used a 2x2 kernel, 'same' padding, and ReLU activation, followed by max-pooling to reduce spatial dimensions. The output was flattened and passed through fully connected layers (512 units with ReLU, 50 units with tanh). The final output was a softmax-activated dense layer for classification. 
    
    \item \textit{DatRet}. DatRet is a deep fully connected neural network for complex data classification. It consists of dense layers with decreasing units to extract patterns from the input. Starting with 512 units and ELU activation, followed by layers with 256, 128, 64, 32, 16, 8, and 4 units, all using ELU activation to capture complex relationships. The final layer is softmax-activated, corresponding to the target classes for classification. 
    
    \item \textit{Transformer}. This neural network, widely used in NLP, leverages self-attention to capture long-range dependencies. Our implementation includes an embedding layer, multi-head self-attention, and feed-forward layers with residual connections and layer normalization. Positional encodings maintain sequence order, and a softmax-activated fully connected layer produces class probabilities, making it effective for sequence classification and language modeling.

\end{itemize}

\subsection{Quality Evaluation}\label{subsec:quality-evaluation}

To assess model quality, we used the standard classification metrics: accuracy for balanced datasets, F1-score (macro-averaged) for imbalanced multi-class dataset, Area Under the Curve (AUC) for imbalanced binary datasets. The evaluation results, conducted with 20 nodes, are shown in Table \ref{tab3}. Our analysis shows that TL consistently matches the results of CL, whereas other DL methods exhibit a noticeable drop in quality.

\begin{table*}[!t]
    \centering
    \caption{Quality results on public datasets. Mean and standard deviation over 20 runs each with different random seeds.\label{tab3}}
    \begin{tabular*}{\textwidth}{@{\extracolsep\fill}llllllll@{\extracolsep\fill}}
        \toprule
        \multirow{2}{*}{\textbf{Dataset}} &
        \multirow{2}{*}{\textbf{Metrics}} &
        \multicolumn{6}{c}{\textbf{Method}} \\ \cline{3-8}
        & & \textbf{CL} & \textbf{TL} & \textbf{FL} & \textbf{SL} & \textbf{SL+} & \textbf{SFL} \\
        \midrule
        MNIST& Accuracy& 99.44±0.04& 99.42±0.06& \textbf{99.51±0.03}& 98.63±0.29& 98.68±0.12&99.36±0.04\\ 
        CIFAR-10& Accuracy& 71.15±0.54& \textbf{70.97±0.57}& 57.71±0.79& 62.27±1.09& 62.84±0.85& 63.12±0.80\\ 
        NICO& F1& 38.62±0.93& \textbf{38.51±1.32}& 01.84±0.66& 36.57±0.90& 37.00±0.56& 37.45±1.51\\ 
        MIMIC-IV& AUC& 88.33±0.05& \textbf{87.96±0.09}& 84.08±0.24& 56.31±4.58& 62.06±5.93& 63.68±5.61\\ 
        BANK& AUC& 90.88±0.23& \textbf{90.87±0.20}& 86.33±0.11& 76.51±5.12& 72.33±8.28& 76.58±8.91\\ 
        IMDB& AUC& 89.92±0.37& \textbf{88.45±0.52}& 85.12±0.45& 84.78±0.60& 84.10±0.65& 85.85±0.55\\
        \bottomrule
    \end{tabular*}
    \begin{tablenotes}
      \item * The best results among the distributed learning methods are highlighted in bold.
    \end{tablenotes}
\end{table*}

We observed that ResNet-18 tends to achieve high accuracy and satisfactory quality consistency across all DL methods. MNIST is a relatively simple and well-structured dataset with grayscale images of digits, making it easier for deep learning models like ResNet-18 to learn discriminative features effectively.

CIFAR-10, however, presents more complex and varied images compared to MNIST, requiring deeper and more sophisticated models for effective feature extraction and classification. Due to the diversity of images, different nodes receive subsets with varying characteristics, such as class distributions, backgrounds, and complexities. We found that FL and SFL struggled to aggregate model updates effectively, as updates from nodes with differing data characteristics did not generalize well to the entire dataset. Sequential learning methods were also less effective for handling complex and diverse features that need to be learned simultaneously for optimal performance. TL, however, closed the performance gap, achieving an accuracy of 70.97\%, representing a 13\% improvement over FL and 8\% over SL.

Compared to these IID datasets, the non-IID NICO dataset posed a significantly higher challenge, even for classic CL. NICO simulates real-world scenarios where data distributions are highly non-IID. FL, in particular, failed to learn from the non-IID data with 20 nodes, as it assumes that data distributions across nodes are similar—an assumption that does not hold in non-IID settings. NICO’s diverse contexts (e.g., dogs on grass vs. dogs on sand) led to imbalanced and diverse data partitions, making it hard for models trained on one node to generalize to another. Sequential learning methods showed better performance than parallel methods in non-IID scenarios, and TL significantly outperformed SL, SL+, and SFL.

DL methods play a crucial role in privacy-sensitive domains like healthcare and finance, where collaboration is required without compromising data privacy. We tested all methods on non-IID datasets, particularly MIMIC-IV (healthcare) and Bank Marketing (financial). Both involve imbalanced binary classification tasks, where one class is underrepresented. SL methods struggled in this context, as they are prone to catastrophic forgetting and skewed intermediate representations, leading to poor performance for the minority class. While FL produced reasonable results, there was still a notable gap compared to CL. TL effectively bridged this gap, achieving an AUC of 87.96\% on MIMIC-IV and 90.87\% on Bank Marketing, representing a 3-4\% improvement over FL.

For text classification on the IMDB dataset, TL demonstrated robust performance as well. IMDB is a balanced binary sentiment analysis task, but the nuances of text data require sophisticated handling of sequential dependencies. While FL and SFL showed declines in AUC (85.12\% and 85.85\%, respectively), TL maintained competitive results, achieving an AUC of 88.45\%, just below CL’s 89.92\%.

The experimental results confirm that TL’s ability to traverse nodes while consolidating BP in the orchestrator is well-suited for handling both structured and unstructured data, including text.

\subsection{Inference Consistency Validation}\label{subsec:inference-consistency-validation}

Consistency in model inference heavily depends on managing the inherent randomness present during the training process. Random number generation (RNG) plays a critical role in several aspects of machine learning, including data shuffling, weight initialization, and dropout layers. In distributed settings, inconsistencies in randomness across nodes can lead to divergence in the models trained locally and, consequently, in the global model. To address this issue, we implemented controls such as the manual configuration of RNG, disabling randomness during training, and consistency testing through iterative training.

\textbf{Manual configuration of RNG}: To ensure deterministic behavior, we manually configured the RNG by setting specific seed values for all random operations in both TL and CL. This approach allowed us to maintain an identical sequence of random numbers during key training stages such as data shuffling, weight initialization, and other random processes. By using the same seed values across both methods, we enforced consistency in how randomness was introduced, ensuring that TL and CL encountered the same conditions during training.

\textbf{Disabling randomness during training}: In addition to setting fixed RNG seeds, we further reduced sources of randomness by disabling certain training techniques that introduce stochastic behavior. This included turning off data augmentation methods such as random cropping, flipping, and rotation, which are commonly applied to enhance model generalization but can lead to inconsistent results across training runs. Additionally, we disabled dropout layers, which randomly deactivate neurons during training to prevent overfitting. By turning off these features, we ensured that the model architecture and training procedure remained stable across both TL and CL, providing a fair basis for comparison.

\textbf{Consistency testing through iterative training}: Once we had controlled and minimized randomness, we conducted iterative training sessions to test the consistency of inference results between TL and CL. These tests involved 20 runs of training and inference under identical conditions, with the primary goal of ensuring that the results were either identical or showed only minimal variance. In each run, we measured key performance metrics such as accuracy, F1-score, and AUC, and compared the results from TL with those of CL. Our findings showed that TL consistently produced inference results that were highly similar to those of CL, with differences falling within an acceptable range of variance. These findings demonstrate that TL is capable of maintaining stable model performance, even in a DL setting.

The results of our inference consistency validation provide strong evidence that TL is capable of producing results comparable to CL, even in a DL setting. By controlling randomness and synchronizing key training processes across nodes, we demonstrated that TL maintains a high level of consistency in inference across a wide variety of datasets and learning tasks. This consistency is crucial for the reliable deployment of DL models in real-world applications, especially in domains where data privacy and performance stability are critical.

In conclusion, our validation shows that TL not only bridges the gap between DL and centralized learning in terms of classification metrics, but also ensures that inference results remain consistent across both learning paradigms. This makes TL a highly viable and robust solution for distributed deep learning tasks.

\subsection{Learning Runtime Analysis}\label{subsec:learning-runtime-analysis}

\begin{table*}[!t]
    \centering
    \caption{Runtime (s) on public datasets. Mean and standard deviation over 20 runs each with different random seeds.\label{tab4}}
    \begin{tabular*}{\textwidth}{@{\extracolsep\fill}llllll@{\extracolsep\fill}}
        \toprule
        \textbf{Dataset} & \textbf{FL} & \textbf{SL} & \textbf{SL+} & \textbf{SFL} & \textbf{TL} \\
        \midrule
        MNIST & 900 ± 45 & 1300 ± 60 & 1350 ± 50 & 950 ± 40 & 800 ± 30 \\
        CIFAR-10 & 1800 ± 85 & 2500 ± 110 & 2600 ± 100 & 1900 ± 90 & 1650 ± 75 \\
        NICO & 2400 ± 120 & 3500 ± 150 & 3600 ± 140 & 2700 ± 130 & 2200 ± 100 \\
        MIMIC-IV & 3000 ± 150 & 4100 ± 180 & 4200 ± 170 & 3300 ± 160 & 2900 ± 140 \\
        BANK & 2000 ± 100 & 3000 ± 130 & 3100 ± 120 & 2200 ± 110 & 1900 ± 90 \\
        IMDB & 3400 ± 160 & 4500 ± 200 & 4600 ± 180 & 3700 ± 170 & 3200 ± 150 \\
        \bottomrule
    \end{tabular*}
\end{table*}

\begin{figure*}[htbp]
\centerline{\includegraphics[width=1.0\linewidth]{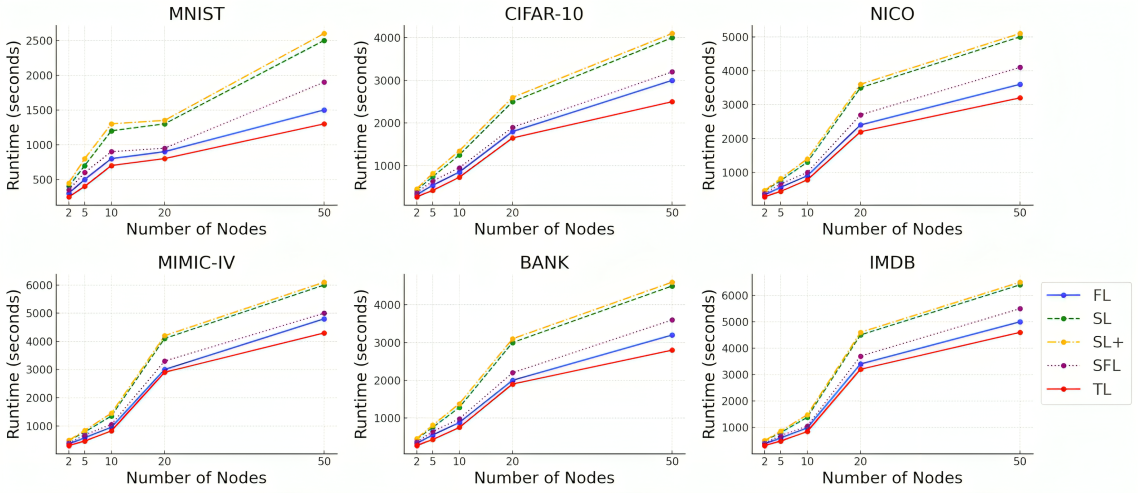}}
\caption{Runtime (in seconds) scalability on public datasets as the number of nodes increases.}
\label{fig3}
\end{figure*}

The runtime of DL methods is influenced by factors such as computation time, communication overhead, synchronization, and the ability to parallelize client operations. Below, we summarize the runtime results for each learning method — FL, SL, SL+, SFL, and TL — based on simulations on multiple datasets and models.

In FL, each client independently trains a model and sends updates to a central server. The total runtime depends on the slowest client (straggler effect), communication overhead, and server-side aggregation. In SL, the model is split between the client and the server. The client handles FP, while the server handles BP. The process is sequential, and the communication overhead for both sending activations to the server and receiving gradients. SL+ avoids label sharing, which increases client-side computation but retains the sequential communication pattern. SFL combines elements of both FL and SL, reducing communication costs by having clients train parts of the model in parallel and aggregate updates on the server. In TL, FP is distributed across clients, while BP is centralized on the orchestrator. Only first-layer activations, first-layer gradients, and last-layer gradients are transmitted, significantly reducing communication overhead compared to transferring full model updates or activations and gradients from all layers.

\begin{equation}
T_{\text{FL}} = \max(T_{\text{comp, client}}) + T_{\text{comm}} + T_{\text{agg}}
\end{equation}
\begin{equation}
T_{\text{SL}} = T_{\text{comp, client}} + 2T_{\text{comm}} + T_{\text{comp, server}}
\end{equation}
\begin{equation}
T_{\text{SL+}} = T_{\text{comp, client}}^{\text{(more layers)}} + 2T_{\text{comm}} + T_{\text{comp, server}}
\end{equation}
\begin{equation}
T_{\text{SFL}} = \max(T_{\text{comp, client}} + T_{\text{comm}}) + T_{\text{agg}}
\end{equation}
\begin{equation}
T_{\text{TL}} = \max(T_{\text{comp, client}}) + T_{\text{comm}} + T_{\text{comp, server}}
\end{equation}

Table \ref{tab4} summarizes the runtime (in seconds) for all methods across various datasets and models, using 20 nodes in the simulation environment. TL consistently achieves the lowest runtime across all datasets and models due to its efficient orchestration of FP and centralized BP. TL minimizes communication overhead by transferring only first-layer activations, making it highly scalable with increasing numbers of nodes. FL and SFL benefit from parallelism, but their reliance on full or partial model updates results in higher communication costs, especially in non-IID data scenarios. SFL generally performs better than FL due to its hybrid approach, which reduces communication delays compared to full model updates in FL. SL and SL+ exhibit the highest runtime. Both methods suffer from sequential processing, which results in significant communication overhead and delays between client and server. SL+ incurs additional computation on the client side, slightly increasing the overall runtime compared to SL.

Figure \ref{fig3} illustrates the runtime scalability of the DL methods across the various datasets as the number of nodes increases. Across all datasets, TL consistently demonstrates the lowest runtime, especially as the number of nodes increases. This indicates its high scalability and efficiency in DL environments. TL is able to minimize communication overhead by performing centralized BP, which makes it highly suitable for large-scale systems with many nodes. FL shows moderate runtime scalability. Its parallel processing helps reduce overall runtime, but the communication and aggregation costs increase as the number of nodes grows, resulting in longer runtime. This is particularly evident in non-IID datasets like NICO and MIMIC-IV, where the disparity in data distributions across nodes amplifies the communication burden. SL and SL+ exhibit significantly higher runtime across all datasets. This is primarily due to the sequential communication between clients and the server during FP and BP. The sequential updates, especially in non-IID settings, further exacerbate the communication delays, leading to sub-optimal performance as the number of nodes increases. SL+ incurs slightly higher runtime than SL due to additional client-side computation, which introduces extra computation and communication overhead. SFL, which combines aspects of FL and SL, performs better than SL/SL+ but still experiences increased runtime as nodes increase. This is due to the partial model updates and aggregation steps, which contribute to the overall communication costs. However, SFL still scales better than SL and SL+ in most cases. Datasets like NICO and MIMIC-IV showcase the challenges of non-IID data in DL. For FL and SFL, the aggregation of models trained on heterogeneous data results in slower convergence and increased communication time. SL and SL+ are particularly sensitive to non-IID data due to the sequential nature of updates, which results in longer runtime as the number of nodes increases. The overall trend highlights that TL is the most scalable and efficient method across both IID and non-IID datasets. In contrast, SL and SL+ are highly sensitive to node count increases due to their sequential communication requirements, making them less suitable for large-scale distributed environments.
    \section{Discussions}\label{sec:discussion}

To provide a comprehensive understanding of the capabilities and trade-offs among DL frameworks, we present a comparative analysis of TL, FL, SL, SL+, and SFL. Table \ref{tab5} highlights key distinctions across critical dimensions such as FP and BP, synchronization mechanisms, communication overhead, scalability, handling of non-IID data, privacy preservation, model quality, and latency sensitivity. The analysis underscores TL’s unique strengths in achieving centralized synchronization for BP while maintaining distributed BP, which reduces communication overhead and enhances scalability. TL excels in handling non-IID data, ensuring high model quality without compromising privacy. In contrast, FL offers scalability and moderate communication efficiency but struggles with non-IID data distributions. Sequential approaches like SL and SL+ prioritize privacy preservation but are limited by high latency and low scalability due to their sequential nature. SFL combines elements of FL and SL, achieving better scalability but at the cost of higher communication overhead. By emphasizing these differences, the table highlights TL’s ability to address challenges inherent in existing DL methods, making it well-suited for scenarios requiring efficiency, scalability, and robust privacy protection, while also framing the need for further advancements explored in this work.

Additionally, we propose several advanced optimization techniques for future research and provide a detailed exploration of potential security ramifications associated with their implementation.

\begin{table*}[!t]
    \centering
    \caption{Comparative analysis of distributed learning frameworks.\label{tab5}}
    \begin{tabular*}{\textwidth}{@{\extracolsep\fill}lllll@{\extracolsep\fill}}
        \toprule
        \textbf{Feature} & \textbf{TL} & \textbf{FL} & \textbf{SL and SL+} & \textbf{SFL} \\
        \midrule
        Forward Propagation & Distributed & Local & Client-side only  & Parallel \\
        Backward Propagation & Centralized & Local & Server-side only & Aggregated \\
        Synchronization & Centralized & Decentralized & Sequential & Decentralized \\
        Communication Overhead & Low & Moderate & High & Moderate \\
        Scalability & High & High & Low & High \\
        Handling Non-IID Data & Strong & Weak & Moderate & Moderate \\
        Privacy Preservation & Strong & Moderate & Strong & Strong \\
        Model Quality & High & Moderate & Low & Moderate \\
        Latency Sensitivity & Low & Moderate & High & Moderate \\
        \bottomrule
    \end{tabular*}
\end{table*}

\subsection{Partial Parameter Update Transfer}\label{subsec:partial-parameter-update-transfer}

As model sizes increase, transmitting updated parameters to all nodes becomes burdensome, leading to substantial I/O overhead and communication latency \cite{ren2019survey}. This issue is particularly pronounced in DL frameworks where frequent updates are required to ensure synchronization across nodes. TL addresses this challenge through the implementation of \textit{partial parameter update transfer}, which significantly reduces the volume of transmitted data by selectively updating only critical components of the model. During training, particularly when techniques like dropout are applied, many connections in the model are temporarily deactivated \cite{dun2023efficient}. Rather than broadcasting updates for all parameters, TL focuses on transmitting only the weights and biases of active connections—those actively contributing to the model's computation during the current iteration. This approach ensures that only essential updates are communicated, thereby reducing the overall data transfer requirements. This selective communication strategy alleviates the load on the system, enabling scalability without compromising model performance. It is especially beneficial in environments with constrained network bandwidth or computational resources. By minimizing unnecessary communication overhead, TL enhances the efficiency of DL, making it more practical for deployment in large-scale systems with numerous nodes.

Additionally, the reduction in communication not only improves runtime but also lowers energy consumption, which is an important consideration in sustainable AI practices. This aspect is particularly relevant for edge computing scenarios, where devices have limited power and processing capabilities. Similar techniques have shown success in prior studies. For instance, \cite{xue2023fedbiad,wen2022federated} demonstrated that selectively updating parameters can enhance communication efficiency without degrading model accuracy. These studies validate the feasibility of TL’s approach and highlight its potential to address scalability challenges in modern DL systems. Looking ahead, the concept of partial parameter update transfer could be further refined by integrating adaptive mechanisms. These mechanisms could dynamically determine which parameters to update based on their contribution to model performance, ensuring even greater efficiency and effectiveness. Additionally, combining this strategy with advanced compression techniques could further minimize data transfer while maintaining the integrity of the updates, pushing the boundaries of DL systems in both resource-intensive and resource-constrained settings.

\subsection{Network Bandwidth Consumption}\label{subsec:network-bandwidth-consumption}

Efficient bandwidth utilization is critical for DL systems, especially in environments with limited network resources or when dealing with large-scale models. TL incorporates several innovative techniques to optimize communication and minimize bandwidth usage during training. 

Caching is a key technique employed to reduce redundant communication during BP. When certain parts of the neural network remain unchanged across training batches, retransmitting the same parameters repeatedly becomes unnecessary. To address this, the orchestrator can cache trainable parameters from the first batch and subsequently collect only minimal information, such as small loss values or gradient updates, from subsequent batches to perform BP. This strategy significantly reduces communication overhead and ensures efficient bandwidth utilization. This approach is particularly effective during fine-tuning phases, where layers of the model are frozen to prevent changes. For instance, in training large models like LLaMA2 \cite{touvron2023llama}, frozen layers are commonly used to preserve learned features while optimizing only specific layers. By avoiding the transmission of redundant parameters for frozen layers, TL effectively minimizes bandwidth consumption, optimizing communication without sacrificing model performance.

In addition to caching, TL leverages activation value compression to further reduce the size of transmitted data. By compressing activations and gradients, TL ensures that only the most critical information is communicated. This strategy is particularly beneficial in scenarios involving geographically distributed nodes or high-latency environments, where excessive data transfer can impede system performance. TL also integrates adaptive synchronization policies to conserve bandwidth in challenging network conditions. For example, nodes can perform multiple FP passes locally before synchronizing with the orchestrator. This reduces the frequency of communication, balancing bandwidth efficiency with training performance. Such adaptive methods enable TL to function effectively in diverse conditions, including intermittent or constrained network environments. By employing these bandwidth optimization strategies, TL achieves a robust balance between communication efficiency and model accuracy. These methods not only enhance scalability and runtime performance but also contribute to sustainable AI practices by reducing energy consumption associated with data transfer. Future work could explore advanced compression algorithms, dynamic caching mechanisms, and network-aware scheduling to further refine bandwidth efficiency and ensure adaptability across a wide range of DL scenarios.

\subsection{Security Implications}\label{subsec:security-implications}

In TL, security is a major concern due to threats posed by malicious orchestrators and malicious nodes. The orchestrator has access to intermediate states, such as activation and gradients during BP, while nodes might try to disrupt the learning process by poisoning their data. An additional risk arises during the index range sharing phase in virtual batch creation, where sensitive information could be unintentionally exposed.

A malicious orchestrator could attempt to infer sensitive input data from the transmitted information, including activations and gradients. Although these values represent abstract, lower-dimensional mappings of the data after several transformations, a determined orchestrator might still attempt to reverse-engineer inputs. To counter this, several approaches can be implemented: (1) Using non-invertible activation functions such as ReLU, which is non-invertible over the range \((- \infty, a)\), where \(a\) shifts dynamically due to bias optimization during BP \cite{dong2017dropping}. This makes recovering input values significantly more difficult. (2) The dropout mechanism, which is commonly used to prevent overfitting, also serves to enhance privacy by randomly deactivating neurons during training \cite{dong2017dropping}. This obscures the relationship between input data and activations, making it harder for an orchestrator to infer the original inputs. (3) Inferring encoded inputs through undisclosed auto-encoders modeled privately on each node, an approach related to the method discussed in \cite{azizian2024privacy}, becomes nearly impossible for the orchestrator even with invertible functions. However, this approach can introduce reconstruction loss. (4) A more robust solution involves encapsulating the activation values in a container created by each node~\cite{seo2024self}. The node-owned containers are transferred to the orchestrator and form \textit{trusted execution environment} on orchestrator's premise. The orchestrator iteratively enters model into the containers for BP in the preset order. Node-owned containers returns the updated model through secure interface prohibiting the orchestrator from taking out the activation values. However, this approach incurs additional cost of transferring containers from nodes to the container.

On the other hand, malicious nodes may attempt to compromise the learning process by submitting tampered data or updates to skew the global model, a challenge that is not unique to TL but is a general concern across all DL approaches. To address this, it is critical to ensure the integrity of the learning process by validating the distribution of data on remote nodes without directly accessing it. (1) Multi-party computation (MPC) \cite{abadi1990secure} can be employed to prevent a single node from manipulating its data undetected. MPC allows nodes to collaboratively verify the distribution of each other's data without exposing the actual datasets, ensuring that remote nodes possess valid data. (2) Knowledge distillation methods can be applied to cross-check data updates across nodes, identifying suspicious or inconsistent updates, and reducing the risk of poisoning attacks. These strategies help ensure that even when data cannot be directly accessed, the global model remains robust against malicious attempts to corrupt the learning process.

During the virtual batch creation phase, nodes share index ranges with the orchestrator to form training batches. While the raw data itself is not shared, these index ranges could still reveal sensitive information about the structure or size of the data on each node. This might allow a malicious orchestrator to infer patterns such as homogeneity or clustering. Several strategies can mitigate these risks: assigning non-sequential, unique values introduces randomness into the index ranges, breaking the correlation between the data and the ranges, while differential privacy adds noise to further obscure meaningful data properties.
    \section{Conclusion}\label{sec:conclusion}

TL introduces a groundbreaking framework that seamlessly integrates CL principles into a distributed architecture, addressing critical challenges in existing DL methodologies. By optimizing FP across nodes and consolidating BP on a central orchestrator, TL achieves high accuracy, scalability, and efficiency. The framework demonstrates robust performance across diverse datasets, including those with IID and non-IID distributions, outperforming traditional methods in preserving data privacy without compromising model quality.

TL’s ability to minimize communication overhead and maintain synchronization highlights its potential as a versatile solution for privacy-sensitive applications, such as healthcare and finance. The evaluation confirms its superior performance in comparison to FL, SL, and SFL, offering significant improvements in accuracy, runtime, and scalability. Future work can further enhance TL’s capabilities by incorporating advanced security measures and optimizing communication strategies, paving the way for its adoption in large-scale, real-world applications. TL sets a new benchmark in DL, bridging the gap between centralized and decentralized approaches with a focus on performance and privacy.
\endgroup


\bibliographystyle{unsrt}
\bibliography{Sections/reference.bib}

\end{document}